# A robotic leg inspired from an insect leg

P.Thanh Tran-Ngoc[1], Leslie Ziqi Lim [1], Jia Hui Gan[1], Hong Wang[2], T.Thang Vo-Doan[3,*], and Hirotaka Sato[1,*]

[1] School of Mechanical and Aerospace Engineering, Nanyang Technological University, Singapore, Singapore.
[2] School of Mechanical Engineering and Automation, Harbin Institute of Technology, Shenzhen, Shenzhen, China.
[3] Institute of Biology I, University of Freiburg, Germany.

**Abstract**

While most insect-inspired robots come with a simple tarsus such as a hemispherical foot tip, insect legs have complex tarsal structures and claws, which enable them to walk on complex terrain. Their sharp claws can smoothly attach and detach on plant surfaces by actuating a single muscle. Thus, installing insect-inspired tarsus on legged robots would improve their locomotion on complex terrain. This paper shows that the tendon-driven ball-socket structure provides the tarsus both flexibility and rigidity, which is necessary for the beetle to walk on a complex substrate such as a mesh surface. Disabling the tarsus' rigidity by removing the socket and elastic membrane of a tarsal joint, the claws could not attach to the mesh securely. Meanwhile, the beetle struggled to draw the claws out of the substrate when we turned the tarsus rigid by tubing. We then developed a cable-driven bio-inspired tarsus structure to validate the function of the tarsus as well as to show its potential application in the legged robot. With the tarsus, the robotic leg was able to attach and retract smoothly from the mesh substrate when performing a walking cycle.

**1. Introduction**

Although traditional robots frequently employ wheels for movement due to their energy efficiency and simple control strategy, this type of locomotion is ineffective when traversing rugged and uneven terrain [1]. Nature has been engineering legged systems through millions of years of animal evolution to enable them to adapt to various terrain. Such diversity in animal locomotion is a great source of inspiration for engineers to develop bio-inspired walking robots during the past few decades. Walking machines or legged robots were designed to move on terrain that are inaccessible to wheeled robots. The capacity of walking machines to change their posture to overcome terrain discontinuities is their most notable attribute [2]. A series of four-legged (quadruped), six-legged (hexapods), eight-legged (octopods) robots inspired by insects and animals has been developed with various structures and functions [3-11]. As the number of legs increases, the robot becomes more stable, but the control system becomes more complex to construct [12]. Although the mechanism and control scheme of legged robots are more complex than those of wheeled robots, the legged robots provide the capability to adapt to complex terrain, which is lacking in the wheeled counterpart [13].

Insects' structure and functions have long been sources of inspiration for developing legged robots due to their high maneuverability and ability to adapt to uneven terrain quickly. For example, HECTOR [14] and LAURON [15] are inspired by stick insects, AMOS [16] is based on cockroaches, and BILL-ANT [17] is inspired by an ant. Feet with different shapes have been designed for some walking robots to improve their locomotive ability in complex terrain [18-21]. While these robots replicated the insect leg structure [22], including the coxa, trochanter, and tibia parts to reinforce their equilibrium while moving, the tarsi were often simplified to a hemispherical foot tip. Although there are attempts to add artificial tarsus and attachment devices to the robotic legs, the applications are limited to flat surfaces [23-27].

Insects can attach to a wide variety of terrain using different attachment systems, which can be broadly classified into smooth adhesive pads, hairy pads, and claws [28-32]. Smooth adhesive pads are generally soft and deformable and can be further subcategorized (e.g., arolia, pulvilli, and euplantulae) depending on their positions on the tarsus [28, 33]. They can adhere well onto smooth surfaces via adhesive forces generated through capillary interactions [34, 35]. On rough surfaces, insects use their hairy pads via attachment forces generated by van der Waals forces [36, 37] and capillary forces [38-40]. Moreover, to ensure the most robust possible grip, both smooth and hairy pads maximize the contact area between the pad and the surface [41]. In cases where this is unfeasible, such as on surfaces with protrusions with mean radii larger than the diameter of the claw tips, insects rely on the claws as means of attachment, as the surface is rough enough that the claws can reliably attach via frictional self-lock [42, 43, 44]. Inspired by the insect attachment systems, RATNIC robot [45] is able to move on vertical pipes thanks to the adhesive pads with enhanced frictional properties that mimic the material of smooth pads. However, this robot only climbed on smooth surfaces. For rough or mesh surfaces, DIGbot [21] includes a foot that mimics insects' claws. The springed tarsus joints allow the claw to change its angle when it contacts the ground and return it to the original position when its grip is released. This passive mechanism helps the robot climb on a screen mesh with a gap of 2cm. However, when the leg angle changes are substantial during turning, spine slippage is more common. Meanwhile, insects can walk smoothly on rough surfaces, including mesh substrates even when they have claws of different shapes and stiffness [42, 46, 47]. For example, beetles with sharp claws put on a piece of wool cloth can walk smoothly without any signs of difficulty.

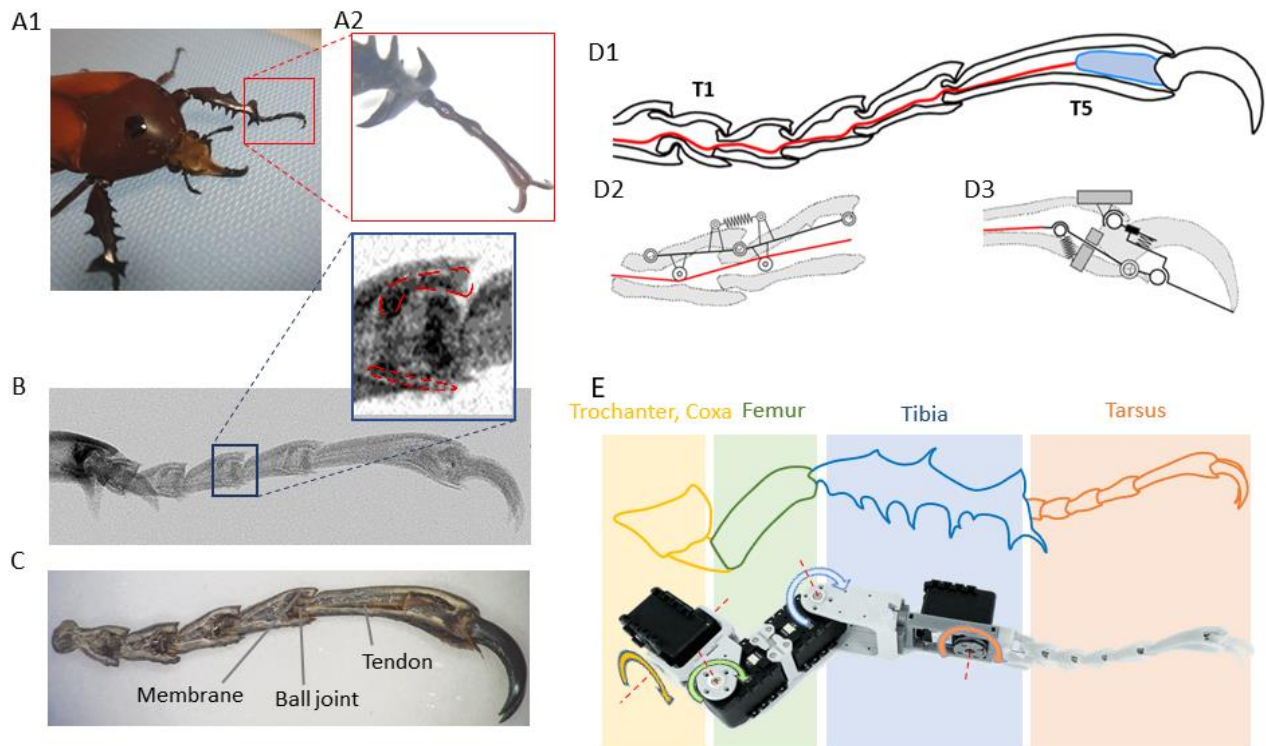

**Figure 1. (A1)** Flower beetle on the mesh substrate **(A2)** the tarsus of the forelegs. **(B&C)** Cross-section image of the tarsus. **(D1)** Diagram of the tarsal chain. **(D2)** Kinematic model of the joint between tarsomeres. **(D3)** Kinematic model of the claw articulation. **(E)** Based on the natural shape of the tarsus, a 3D printed tarsus was designed and manufactured, and this structure was connected to four actuators to replicate the movement of the foreleg [11].

This paper shows that the tarsus' ball and socket structure is responsible for walking adaptation on different terrain in *Mecynorhina torquata* beetle (Figure 1A-C). The tarsus' flexibility, the result of a chain of ball and socket structure that links the tarsal segments, suggests that it will bend during walking, allowing the claw to be passively angled into a position where it can detach when the leg swings forward. Moreover, the tarsus is actuated only via a single muscle, the claw retractor muscle, within the tibia and the tarsal promotor muscle located at the end of the basitarsus [22, 48]. When actuated, the tarsus was bent and stiff while the claws were opened to grasp the substrates (Figure 1D1, D2 and D3). The kinematics model of the tarsus is simplified to a chain of linkages with elastic joints actuated by a cable-driven mechanism. The claws are simplified based on the unguitractor apparatus, which links the tendon and the claws together [42, 49]. On the basis of the mechanism of the beetle's claw (Figure1D3), we designed a new mechanism for the artificial tarsus that made the claws open and rotate downwards when pulling the string, as well as return to its original position when released (Figure 6A2). We then built a bio-inspired tarsus based on the structure of the beetle tarsus, which was able to replicate the function of a real tarsus when attached to a robotic leg (Figure 1E).

## 2. Materials and Methods

### 2.1 Animals

We used adult flower beetles, *Mecynorhina torquata*, purchased from the Kingdom of Beetle Taiwan Co., for the experiments. The beetles were kept in separate cages (15 cm × 15 cm × 20 cm) in a rearing system (NexGen IVC system) and fed with sugar jelly twice per week. The system was maintained at 25°C and relative humidity of approximately 60%. Only male beetles (5–7 cm) were used for all experiments. All experiments were carried out following the guidelines of the National Advisory Committee for Laboratory Animal Research.

### 2.2 Tarsus anatomy

The foreleg with tarsus was removed from a dead beetle and dried before X-ray imaging. The X-ray image of the foreleg tarsus was captured using a micro-CT scanner to show the tarsus structure in different shades of black and white (Figure 1B). For cross-sectional view, sandpaper was used to remove the cuticle from half of the tarsus. The structure of each tarsomere was exposed and captured under the microscope (Figure 1C).

### 2.3 Destruction of tarsus' structure and functions

To eliminate the flexibility of the tarsus when the beetle walked on the mesh surface, we used heat-shrink tubing. In preparation for the tubing process, the beetle was restrained on a frame using nylon cable ties wrapped around the body and legs. Using a cylindrical tube with a diameter of 2 mm, we covered the tarsus and a portion of the tibia (Figure 2C). Because this tube has a shrink temperature of approximately 90°C, a soldering iron of approximately 150°C was used as a heat source. The heated metal tip was moved along the tubing within 20 s to shrink and wrap it tightly around the tarsomeres, as well as the joint between the tarsus and tibia. After the experiment, these tubes were removed by a pair of scissors, and the beetles were returned to the rearing system. After being allowed to rest for one day, the beetles were allowed to walk on the mesh substrate. The stepping patterns were measured to determine whether the heat source from the soldering iron impacted the tarsal performance.

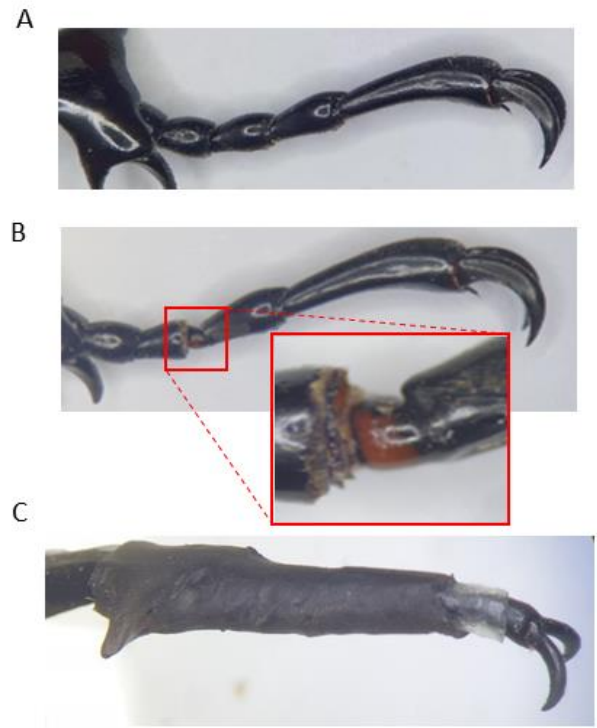

**Figure 2.** Physical constraints in the rigidity and flexibility of tarsus: **(A)** the intact tarsus, **(B)** after cutting the membrane connecting the third tarsomere, **(C)** after tubing the tarsus.

To remove the stiffness of the tarsus, the cuticle connecting the third tarsomere to the second tarsomere and the membrane underneath it was removed using a pair of scissors to decrease the rigidity of the tarsus. A pair of micro-dissecting spring scissors (Vannas tweezer, pattern no. 5, tip size 0.05 × 0.01 mm) was used to remove the cuticle around the tarsomere (Figure 2B) until the membrane was exposed. The membrane underneath the cuticle was then cut.

### 2.4 3D model and fabrication of the bio-inspired tarsus

Based on the tarsomeres' structure, we designed a 3D model of the artificial tarsus using a CAD software (SOLIDWORKS). The prototype was then 3D printed by ANYCUBIC Photon mono printer using Acrylonitrile Butadiene Styrene material. The shapes of tarsal articulations are arranged in such a way that the tarsal chain bends downwards. A string and steel alloy compression springs (outer diameter of 4.5 mm, spring constant of 0.54 N/mm) were used to replicate the function of the tendon and the membrane of the beetle's tarsus, respectively (Figure 6B).

### 2.5 Measure the vertical and horizontal force when tarsus bending.

To measure the vertical and horizontal forces when the artificial tarsus bend in the rigid condition, we used the digital force gauge SAUTER FH 100 (range 100N, resolution 0.05N). The force gauge was mounted perpendicularly on the basement for vertical force measurement. However, the force gauge was placed on the slide block of the screw guide rail to measure the hooking force.

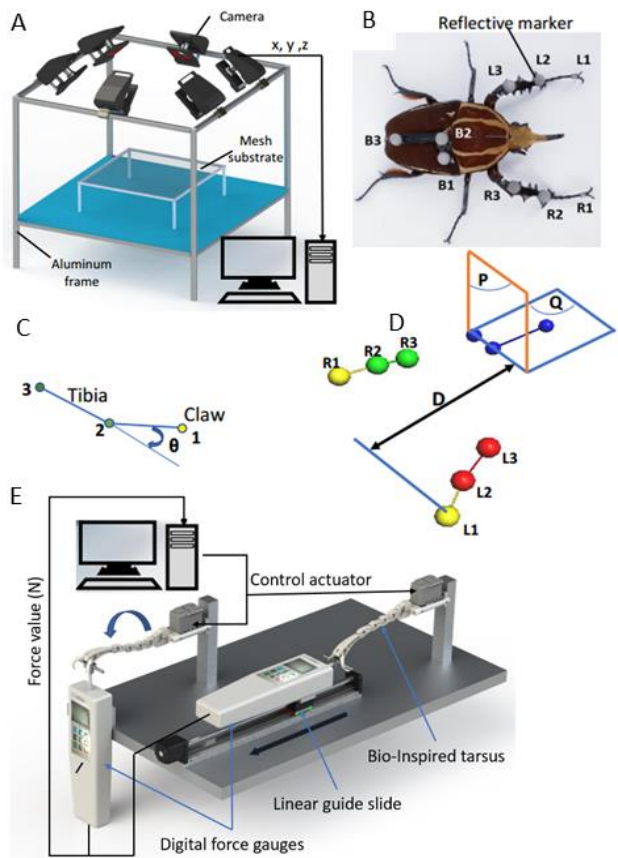

**Figure 3.** Set up for the experiment to track the trajectories of the beetle legs and body. **(A)** The 3D motion-capturing system consists of (i) six T40 VICON cameras (Vicon Motion Systems, Oxford, UK), each with an equal resolution of 4 megapixels (2336 × 1728) for tracking the position (100 fps), (ii) the VICON server for recording, storing, and showing the 3D position collected by the cameras, and (iii) MATLAB software (MathWorks, Natick, MA, USA) for analyzing the angular displacement and the distance among positions. **(B)** Three reflective markers were stick on each front leg to represent the tarsus and tibia segments. The L-shape structure with three reflective markers was attached to the beetle's body, allowing the reference plane to be made. **(C)** Diagram showing how angular displacement was calculated. The angle between claw and tibia is used to describe the bending ability of the tarsus. **(D)** These markers were displayed by the points on the Nexus software. To present the tibia and tarsus segments of each front leg and the beetle's body, we linked the three points on the left and right front leg and three points on the body. Investigate the movement of the front legs of the beetle when walking on the plate and mesh substrates by measuring the displacement D of claws. **(E)** The setup used to measure the horizontal force (hooking force) and vertical force of artificial tarsus when it switched to rigid condition includes (i) The digital force gauge (SAUTER FH100) to measure force values and transfer data to PC, (ii) the linear guide slide to move tarsus during the hooking force measurement. And (iii) actuator AX12A for pulling the string to bend the tarsus.

The set of artificial tarsus and actuator was fixed on the aluminum frame. This actuator was used to switch between the flexible and rigid conditions of the tarsus. In the rigid condition, the claws will open and bend down to hook on the stain-steel rod with 5mm diameter. This rod was attached to the end of the force sensor (Figure 3E).

The vertical force was produced by pushing the claws on the rod when the tarsus bent down. The maximum force was achieved when the tarsus could not bend anymore or the actuator was overloaded. The hooking force was produced by moving away from the slide block. It reached the maximum force when either the claws slipped out of the rod or the tarsus was destroyed.

*2.6 Tracking the trajectories of legs and body*

We used a hexagonal mesh made from polyester with a gap of 2 mm across and a Styrofoam plate as the surface for beetle walking. A three-dimensional (3D) motion capture system (6x T40s VICON cameras) was used to record the trajectories of the front leg and body at 100 fps when the beetle walked on the mesh surface (Figure 3A). To calculate the angular displacement between the tibia and tarsus segments, three retro-reflective markers (L1 and R1, 2-mm cylinder markers; L2, L3, R2, R3, 3-mm spherical markers) were attached to each front leg to denote the tibia and tarsus segments (Figure 3B and C). To identify the stepping patterns when the beetle is walking, an L-shaped frame with three markers (B1, B2, B3, 3-mm spherical markers) was also attached to the beetle's body to measure the distance between the claws and the reference plane made by the L-shaped frame (Figure 3B and D).

*2.7 Demonstration of the 3D printed tarsus*

To replicate the movement of the beetle's front leg, we used a set of four servo motors (AX12A): three motors for the rotation of the coxa-trochanter, femur, and tibia and one motor for pulling the string to switch the artificial tarsus between rigid and flexible conditions (Figure 1E). The tarsal performance was evaluated by replicating the trajectory of the beetle's tibia when walking on the mesh. The tibia-tarsus joint trajectory of the beetle forelegs during walking was recorded and scaled up eight times to fit the size of the robot leg. The Denavit-Hartenberg (DH) parameters of the robotic leg were then configured in the MATLAB kinematics module (Robotics System Toolbox) to translate the scaled trajectory into the motors' joint angles (Figure 10B). These parameters were transferred to the robotic leg to reproduce the trajectory of the tibia-tarsus joint of the beetle.

A nylon square mesh with 25 mm spacing was used as a substrate to evaluate the performance of the 3D printed tarsus. One retro-reflective marker (3-mm spherical marker) was attached to the fifth segment (T5) of the artificial tarsus to track its movement. To visualize the mesh's deformation, we placed four retro-reflective markers (3-mm spherical markers) along the perimeter of the area where the claw contacts the mesh. The displacement of the mesh was calculated by the mean displacement of these four markers. The positions of the markers were tracked by the motion capture system (VICON) at 100 fps.

After collecting the 3D motion data of the five markers, the VICON Nexus software was used to reconstruct the markers' positions and export data in CSV format. A custom MATLAB program was then used to analyse the displacement of the artificial pretarsus and the deformation of the mesh

## 3. Results and discussions

### 3.1 The tarsus bent differently on different terrain

The tarsus is made up of a chain of individual tarsomeres that resemble hollow cylinders, with two consecutive tarsomeres being connected in a ball and socket configuration. In the *M. torquata* beetle, the tarsus consists of five tarsomeres. To verify if the tarsus plays a crucial role in smooth walking, we tracked the forelegs of the beetles with motion capture cameras to determine if they bend during walking and if there are any significant differences in the tarsus when walking on surfaces of different roughness. This was performed by allowing the beetle to walk on two different terrain: a solid, continuous surface (Styrofoam board) and a flexible, mesh surface (Figure S1A and B).

The tarsus bent during walking on both terrain but deformed to a greater degree on the mesh compared to the solid surface because the mesh has deeper depressions or holes (Figure 4A and B). When the claw was on the mesh, the claw penetrated through the surface. When swinging the front legs forward, the claw was not released from the depression immediately, causing the tarsus to deform before the claw was eventually freed. However, such a deformation did not cause any struggle in walking or refrain the beetle from retracting the tarsus out of the mesh. In contrast, on the solid surface, the shallow depressions mean that the claw can be freed more easily when the legs swing forward; hence, the deformation of the tarsus is smaller than that on the mesh. Specifically, the ability of the tarsus to bend is described by the angle between the claw and the tibia, as shown in Figure 5A. The change in bending angles when walking on the mesh and the foam board was significant, which was 55.7° ± 4.4° and 27.7° ± 5.4°, respectively (Figure 5A, t-test,

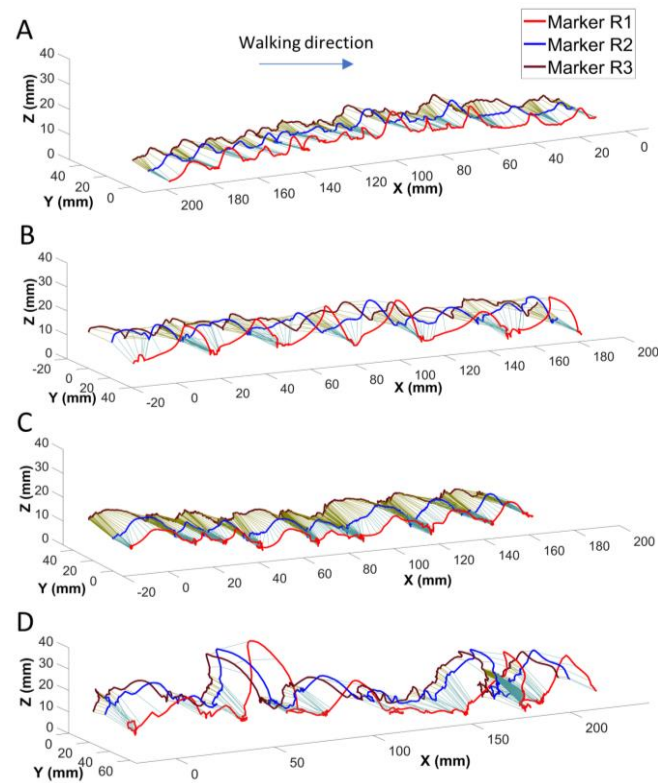

**Figure 4.** The trajectories of beetle right foreleg (marker R3 and R2 were attached on tibia, and marker R1 was attached on claw) were captured in four cases: Beetles with intact tarsus walking on **(A)** Styrofoam plate and **(B)** mesh substrate, Beetles walked on the mesh substrate with **(C)** cutting the membrane of 3$^{rd}$ tarsomere and **(D)** tubing tarsus. The time between two consecutive frames was 50 ms .

N = 5 beetles, P = 9.04E-06, df = 8, Supplementary Table S1). The cycle time needed for the beetle to release its grip, swing its leg forwards, and touch down on the walking surface again was increased by approximately 10% when the beetle was walking on the mesh as compared to on the continuous ground, with a value of approximately 446.1 ± 50.5 ms and 406.6 ± 67.8 ms, respectively (Figure 6C). However, there was an insignificant change in the cycle time between these cases (t-test, N = 5 beetles, P = 0.1631, df = 8, Supplementary Table S2). The beetles did not show any struggle or obstruction when walking on the mesh. Overall, all these suggested that tarsal deformation would affect the speed of the beetle when walking on the mesh, but its flexibility would prevent any blockage when retracting the tarsus out of the mesh.

### 3.2 Rigidity enables secure attachment, passive aiming, and righting of the tarsus and claws

Despite the flexible structure of the tarsus, it always stiffens and is bent inward towards the walking surface when actuated. How can the tarsus increase its stiffness and rotate in such a specific inward direction, when the high flexibility of the ball-socket joints suggests that there should be no directionality in the rotation? While the ball-socket joints provide both flexibility and rigidity to the tarsus, it is the flexible membranes that connect the ball to the socket would be necessary for not only protecting the joint but also to maintain the bending direction of the tarsus. To evaluate the roles of the tarsus' rigidity as well as its passive aiming and righting, we let the beetle to walk on the mesh and compared the bending angle between an intact tarsus and a tarsus with removed socket and cut membrane in the front leg.

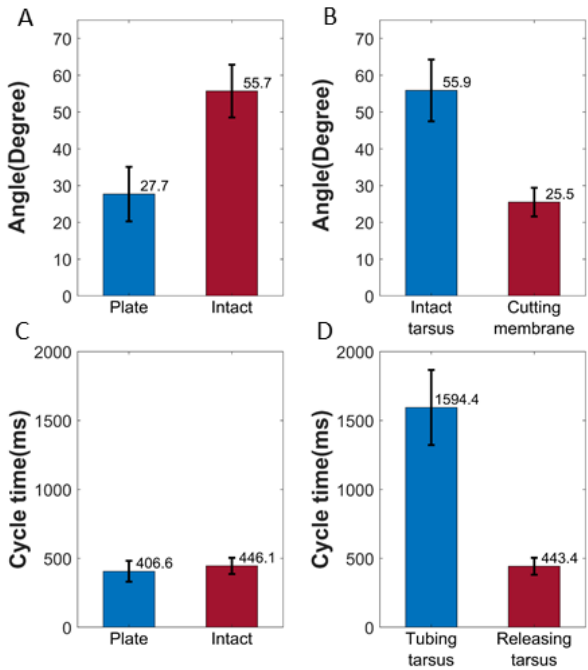

**Figure 5.** The angular displacement of the claw when: **(A)** the beetle is walking on the mesh and Styrofoam plate (N = 5 beetles, n = 25 trials), **(B)** the beetle is with intact tarsi and cut membrane walking on the mesh (N = 3 beetles, n = 15 trials). The cycle time of the beetle foreleg when **(C)** the beetle is walking on the mesh and Styrofoam plate (N = 5 beetles, n = 25 trials), **(D)** the beetle is with intact tarsi and tubed tarsi (N = 5 beetles, n = 25 trials). The error bar is the standard deviation.

The outermost layer of the tarsomere is a hard shell, called the cuticle, and underneath the cuticle is a membrane, which wraps around a tendon (Figure 2B). Within the tarsus, this tendon passes through the hollow path within the tarsomeres. One end of the tendon is connected to the claw retractor muscle within the tibia, whereas the other end is connected to the unguitractor plate and the claws. The tendon lies within the middle plane from both sidewalls and leans close to the inner wall of the tarsus. Such an arrangement would cause the tarsus to always bend inward when the tendon is pulled. When the claw retractor muscle contracts, tension increases within the tendon, causing the tarsomeres to rotate inwards, compress on each other, and thus turn stiff. When reaching a threshold, the tendon pulls open the claws and grip onto the surface [46, 48, 49]. The increased rigidity of the tarsus and the open-angle of the claws then provide strong attachment to the mesh. With the intact tarsus, the claw was firmly anchored to the mesh when activated. When the beetle starts to swing its foreleg forwards, the claw retractor muscle relaxes to remove the tension within the tendon and increase the flexibility of the tarsus. The tarsus was bent because the claw was not completely removed from the surface, causing the angular displacement of the claw relative to the foreleg's tibia to increase. After a sufficient pulling force freed the claw, the foreleg continued its swing, and the claw then came into contact with the mesh and activated again, thereby resetting the claw's angular displacement.

When the membrane of the tarsomere was removed, there was a noticeable change in the trajectory of the claw. In particular, the Z-axis displacement of the claw with the cut membrane was much smaller than that of the claw with the intact tarsus (Figure 4A and C). Besides, the angular displacement of the claw with cut membrane was significantly decreased (t-test, N = 3 beetles, P = 2.42E-05, df = 4, Supplementary Table S3). The mean angle of the intact tarsus was 55.9° ±2.3°, whereas it was only 25.5° ± 1.6° following the removal of the membrane in the front tarsus (Figure 5B). Such an angular displacement of the intact leg was insignificantly different from that of an intact beetle walking on the mesh (55.7° ± 4.4°) (Figure 5A and B, t-test, P = 0.4418, df = 6, Supplementary Table S4), which confirmed that the intact leg was not affected by the removal of the membrane on the opposite leg, and it can operate normally during the experiment. The decrease in bending angle of the tarsus with cut membrane indicated that the claws just leaned on the mesh surface instead of penetrating into it during the stance phase. When the beetle moved forward, the claws only flipped and leaned on the mesh as the foreleg finished the walking cycle. Without a secure attachment of the claws to the mesh, the displacement of the tarsus with the cutting membrane became smaller than that of the intact one (Figure S1C). This result shows that the ball-socket joints play a vital role in turning the tarsus from a flexible to rigid state when it needs a strong attachment to the mesh, and the flexible membrane of the joints enables passive aiming and righting of the tarsus and claws to the walking substrate.

*3.3 Flexibility is necessary for the tarsus to retract from the mesh surface*

Despite the complex structure of the tarsus and claws, the beetles showed only a slight increment in walking step duration of 39.5ms when walking on the mesh as compared to that when walking on a flat surface. Such efficiency would come from its capability to rapidly switch the tarsus from rigid to flexible states when retracting the leg out of the mesh during the swing phase. To evaluate the role of tarsus' flexibility, we compared the recorded trajectories of the claw and the displacement of the claw relative to the thorax between the intact tarsus (Figure 2A) and the tarsus that was fixed by a small tube to inhibit its flexibility (Figure 2C).

In the case of the intact tarsus, the beetle had no difficulty moving on the mesh surface, as shown in its trajectory (Figure 4A), although there is a remaining attachment that pulls the tarsus at the beginning of the swing phase (Figure S1B). In contrast, when the tarsus was fixed, the beetle had significant difficulties in removing its claws from the mesh (Figure 4D), which distorted the regular cycle between each leg stroke significantly. Specifically, the mean cycle time of the beetle foreleg in this experiment is much larger than that of an intact leg, with 1594.4 ± 142.5 ms compared with 446.1 ± 50.5 ms, respectively (Figure 5D, t-test, N = 5 beetles, P = 7.33E-08, df = 8, Supplementary Table S5). When the claw retractor muscle relaxed, tension within the tendon was removed, the membrane helped to return the tarsus to its original shape and flexibility, and thus relieving the attachment of the tarsus to the substrate. Such a property enabled the beetles to retract the tarsus from the mesh during the swing phase. However, when the tarsus was tubed, it could attach to the mesh but was not able to retract as it could not release the strong attachment effectively. When the beetle tried to remove its claws from the mesh, the tarsomeres were unable to bend or rotate to an angle that allowed the claw to detach from the mesh, and the front legs must repeatedly swing until the claw was successfully detached via brute force. This process causes the vibration movement in between each leg swing, as captured in Figure S1D. As such, the flexibility of the tarsus due to its ball and socket configuration is necessary for insects to detach from surfaces.

In addition, we let the beetles walk on the mesh again after removing the tube to ensure the previous heating during fixation did not damage the tarsus. There is no significant difference in cycle time between the beetles after removing the tubes and those with intact tarsus, with a value of approximately 443.4 ± 48.4 ms and 446.1 ± 50.5 ms, respectively (N = 5, n = 25) (Figure 5C and D, t-test, N = 5 beetles, P = 0.4664, df = 8, Supplementary Table S6). This implies that the tarsal performance was not impacted by the heat that was applied to the tube.

*3.4 The bio-inspired tarsus*

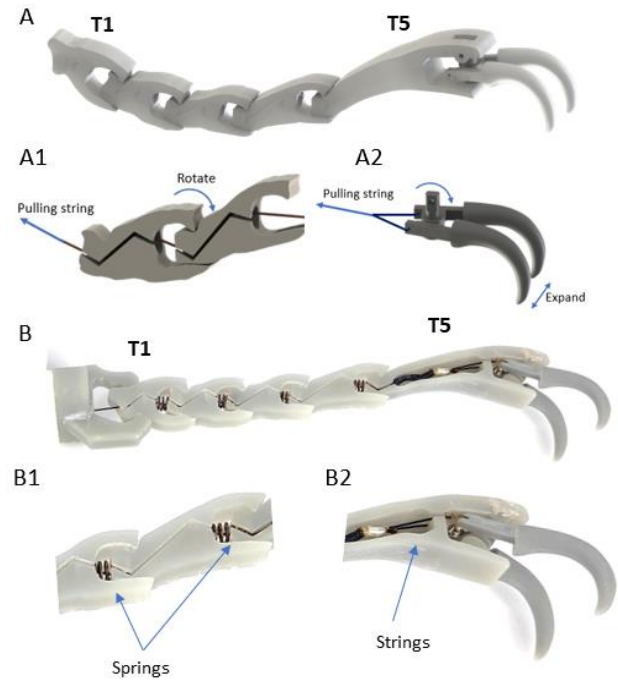

**Figure 6. (A)** the 3D model of the bio-inspired tarsus with **(A1)** the mechanism of tarsal segments made the tarsal segments bend when pulling the string and **(A2)** the claws mechanism made the claw open and rotate downwards when pulling the string. **(B)** The prototype was manufactured by the 3D printer, **(B1)** the tarsal segments were connected by the spring, and **(B2)** the claws were attached to string.

As aforementioned, the ball-socket joints along with the elastic membrane of the tarsus enabled not only strong attachment but also an easy detachment of the claws when the beetles walked on the mesh. Integrating such attachment devices into the robotic legs would enable the robot to walk on the mesh substrate efficiently. Thus, we designed and built a prototype of the bio-inspired tarsus for the robotic leg based on the structure of the beetle's tarsus and the kinematic model of the joints between the tarsomeres and the claws (Figure 6).

The artificial tarsus consists of five tarsomeres connected to each other with simplified ball-socket joints using elastic springs. The claws were attached to the last tarsomere with a revolute joint. A string running through the tarsus connected the claws' bases on one end to the motor on the other end. By pulling the string, the tarsomeres were compressed to each other, which increased the tarsus' stiffness, bent it inward, and opened the claws (Figure 9A2). When the string was released, the tarsus returned to its original position because of the elastic force of the springs (Figure 9A1).

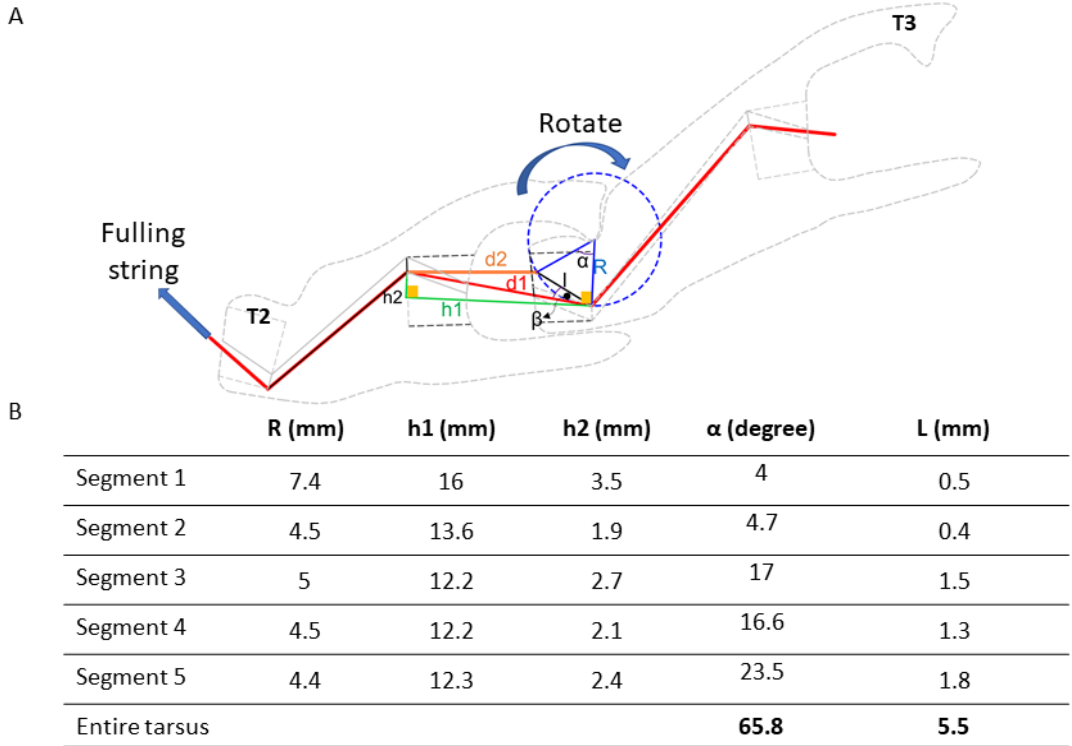

**Figure 7.** Kinematic model of the tarsal segment. **(A)** The diagram shows the relationship between the bending angle of each tarsal segment and the traveling distance when pulling the string. **(B)** The value of variables R, h2, h1, and α for each tarsal segment are measured from the design. The displacement length (L) was calculated based on these variables.

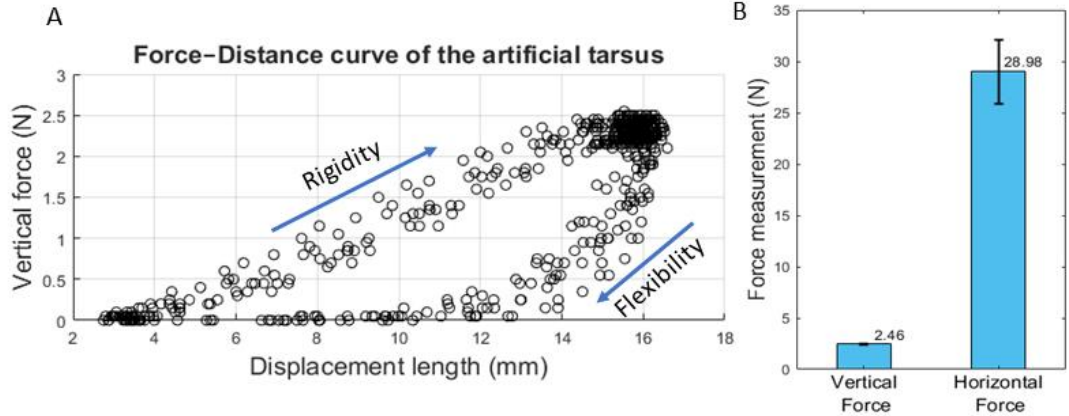

**Figure 8.** Measure the force of the bio-inspired tarsal chain. **(A)** Force–distance curve obtained on the tarsal chain loaded in the vertical direction. **(B)** Force generated by bending artificial tarsus (N = 20 trials for vertical force and N = 5 trials for horizontal force).

The kinematic model for the rotation of the artificial tarsus is shown in Figures 1D and 7A. Figure 8A shows one tarsal segment rotating around the joint when pulling the string and is described by equations 1 to 4.

$$L = d_1 - d_2 \quad (1)$$

$$d_2 = \sqrt{d_1^2 + l^2 - 2 \times d_1 \times l \times \cos(\beta)} \quad (2)$$

$$l = \sqrt{2R^2(1 - \cos(\alpha))} \quad (3)$$

$$\beta = \frac{\alpha}{2} - atan\left(\frac{h_2}{h_1}\right) \quad (4)$$

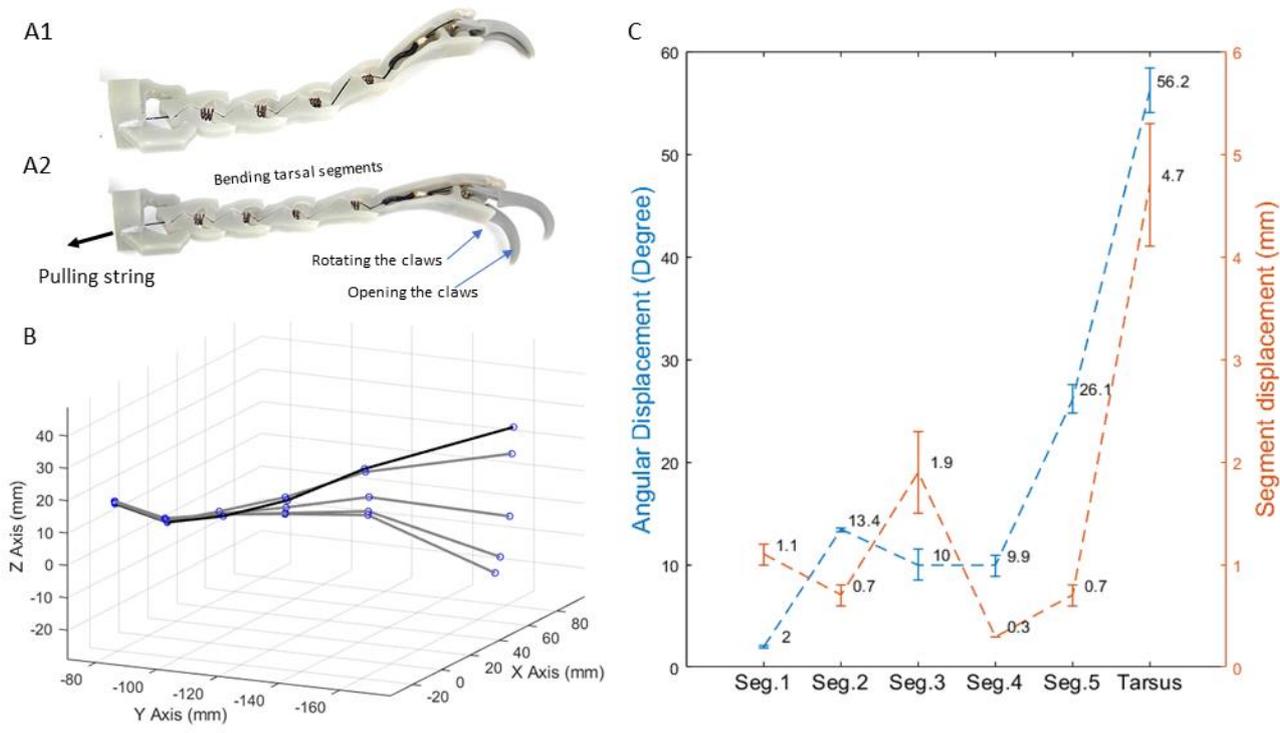

**Figure 9.** The bending ability of the artificial tarsus. **(A)** The shape of the artificial tarsus changes between two conditions: **(A1)** flexible and **(A2)** rigid when the actuator pulls the string within. **(B)** The kinematic range of the artificial tarsus was plotted by capturing the trajectory of 8 reflective markers attached to the artificial tarsus. **(C)** The angular displacement and length displacement of each tarsal segment and the whole tarsus were calculated based on the trajectory of each segment.

Based on the 3D model, the maximum angular displacement from segment 1 to segment 5 was measured (Figure 7B). Equation 1 can be derived to calculate the retractor string displacement between flexible and rigid conditions. The displacement length is 5.5 mm, and the entire tarsus can bend 65.8°.

*3.5 Actuation and stiffness of the bio-inspired tarsus*

When the motor was not actuated, the string was relaxed and the tarsus was bent upward due to the elastic springs while the claws were closed (Figure 9A1). When the motor actuated, it pulled the string replicating the tendon, the artificial tarsus was bent downward, and the claw was opened (Figure 9A2). During the pulling process, the 3D printed tarsus was bent and switched to the rigid condition. The angular displacements of each segment were measured to show the bending ability of the structure. All segments of the structure were controlled so that the entire tarsus bent down. The bending angles of the artificial tarsus were calculated and compared with the theoretical angle ($\alpha$), as shown in Figure 7B. The angular displacement from segment 1 to segment 5 are 2° ± 0.1°, 13.4° ± 0.3°, 10° ± 1.4°, 9.9° ± 0.9°, 26.3° ± 1.5°, respectively. Most of the experimental angles are smaller than the theoretical data except those of segments 2 and 5. While there was a small variation observed in segment 5 (26.3° ± 1.5° from the experiment as compared to 23.5° from the theory), the calculated angular angle from segment 2 is nearly 3 times higher than the analytical result (Figure 9C).

This difference can be explained as the socket structure around the joint of segment 2 was deformed when the actuator continued to pull the string while other segments had reached their limit. Besides, the segments were compressed together during the bending, thus adding the translational degree of freedom to the tarsus. By measuring the length between two joints of the adjacent segments, the displacement length from segment 1 to segment 5 are 1.07 ± 0.1mm, 0.74 ± 0.15mm, 1.9 ± 0.37mm, 0.31 ± 0.04mm, 0.68 ± 0.07mm, respectively (Figure 9C). Therefore, the motor of the tarsus pulled the string for a distance of 13.1 ± 0.4 mm instead of 5.5 mm, as calculated in theory. However, the angular displacement of the entire tarsus could achieve up to 86% of the theoretical value, including the effect of the tarsal weight (56.4° ± 2.2° from the experiment as compared to 65.8% from the theory).

When the bio-inspired tarsus was actuated, it bent downward and pressed the load cell with a maximum force of 2.46 ± 0.06N in the vertical direction. Force–distance curves show that the bio-inspired tasus becomes much stiff if the string is tightened. Moreover, the curve shapes between the rigid and flexible states are different due to the elastic force of the springs (Figure 8A). The actuated artificial tarsus could also produce a maximum hooking force of 28.98 ± 2.46N (N = 5 trials) (Figure 8B). However, the strength of the claws limited the maximum possible hooking force as the bases of the claws were broken at the end. Using other materials would enhance the strength of the claws and thus increase the hooking force.

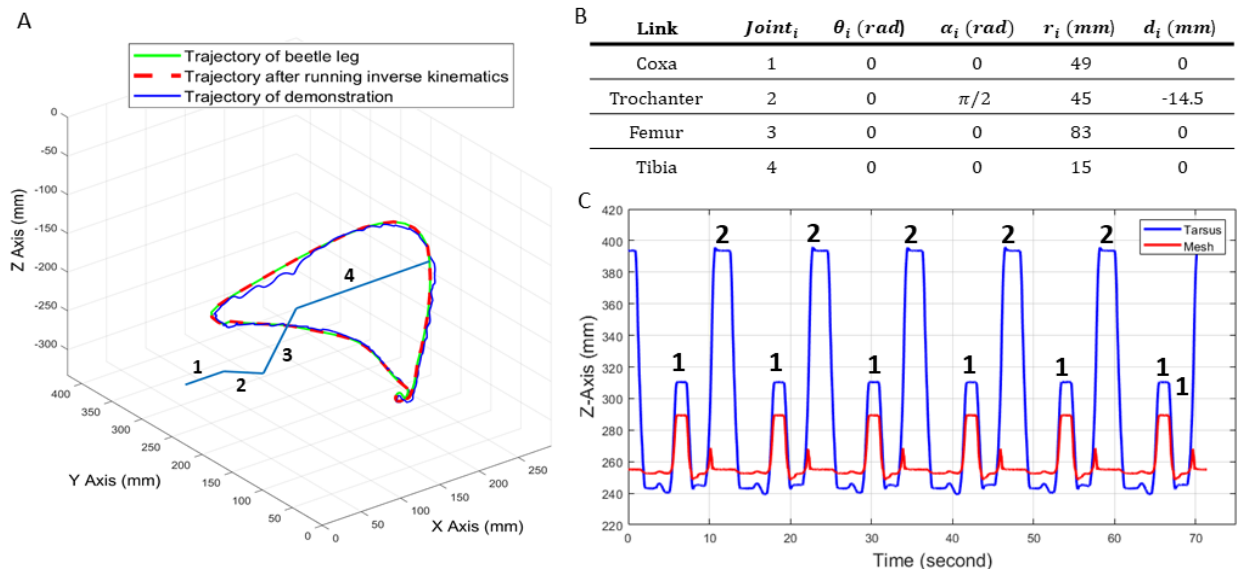

**Figure 10.** Demonstration with the 3D printed tarsus. Visual deformation of the artificial tarsus. **(A)** The inverse kinematics method was applied to control the movement of the robot leg so that the robot tibia can generate a similar trajectory to the beetle's tibia. **(B)** Denavit-Hartenberg parameters for the configuration of the robotic leg: Segments 1, 2, 3, and 4 represent coxa, trochanter, femur, and tibia respectively **(C)** Simulating the movement of the robot leg on the mesh substrate. When the motor "contracts," which is denoted as state 1, the tarsus is bent so the claws attach to the mesh. Moreover, the claws still hook on the mesh when the leg switch to the swing phase, the height of the mesh follows that of the claws closely, indicating attachment. When the leg return and the motor "relax", which is denoted as state 2, the tarsus switch to flexible condition. Then, the claws can come out of the mesh, the height of the mesh does not rise past its resting height, indicating release.

*3.6 Demonstration of the bio-inspired tarsus with robotic leg*

The bio-inspired tarsus was assembled to the tibia end of a robotic leg that replicates the beetle leg. The tarsus interacted with the mesh substrate while the leg reproduced the walking step of the beetle walking on the mesh (Figure 1E).

The inverse kinematics allowed us to mimic the beetle's tibia-tarsus joint trajectory with high accuracy. Thus, the robotic foreleg was able to mimic the same trajectory. The fluctuations were captured in the swing phase due to limitations of the AX-12A as the trajectory was divided into steps, and each servo was updated sequentially (Figure 10A). The Swing and Stand phases are two distinct sections of the trajectory. At the end of the Swing phase, the tarsus switched to the rigid condition, bending the tarsus and opening the claws, allowing the claws to hook onto the mesh. Nearing the end of the Stand phase, the tarsus switched to the flexible condition, allowing the claws to disengage from the mesh for the foreleg to swing freely.

When the motor "contracted" (denoted as state 1, Figure 10C), the prototype tarsus is bent and hooked onto the mesh surface securely. In this state, when the artificial tibia swung upwards, the height of the mesh markers followed that of the claws closely, indicating that the claw was hooked securely. Then, the robotic leg moved back so that the claws return to its original position in the mesh. When the motor "relaxed" (denoted by as 2, Figure 10C), elastic components within the prototype tarsus allowed it to become flexible, and no difficulties were observed when the artificial tibia swung upwards. Similarly, this can be seen by how the height of the mesh markers returned to their original position while the claws continued to increase, indicating that the claw was free from the mesh (Figure 10C). Overall, the effects of the flexibility and rigidity of the tarsus are further verified through this demonstration.

**4. Conclusion**

In this study, we hypothesized and confirmed the role of the tarsal ball and socket configuration and the necessity of the tarsus' flexibility as well as its ability to bend and become rigid in insect walking. In summary, when the tendon is pulled, the ball and socket configuration causes the tarsus to bend at a certain orientation and increase its rigidity. Besides that, the claws rotate inwards and open up. These conditions help insect claws to hook securely into the walking surface. When the tendon is released, the elasticity of the membrane connecting the tarsomeres returns the tarsus to its original position and flexibility. In addition, the claws rotate outwards and close. Such conditions help the beetle to easily retract their claws out of the walking surface. Disabling the tarsus' ability to turn rigid by removing the socket and the joint membrane made the claw unable to aim and hook onto the mesh walking surface securely. In addition, when the tarsus' flexibility was disabled by the tubing, the beetles faced significant difficulties in retracting the claws out of the walking surface.

Finally, we developed a bio-inspired tarsus to not only verify the role of beetle tarsus but also provide the new capability of walking on the mesh to legged robots. Using the beetle tarsus as a reference, we constructed a prototype of a robotic leg with integrated artificial tarsus to replicate the movement of the beetle foreleg. The ball and socket configuration allowed the bio-inspired tarsus to switch between rigid and flexible conditions, thereby allowing it to hook into and detach from the mesh surface without significant difficulties. Such capabilities suggest that the functions of the ball and socket configuration can be replicated using artificial means and incorporated into legged robots to allow them to have the same capabilities as natural beetles.

**Acknowledgments**

This work was supported by the Singapore Ministry of Education (RG140/20) for H.S. T.T.V.-D. is currently supported by Human Frontier Science Program Cross-disciplinary Fellowship. The authors appreciate Mr. Chong Bing Sheng for proofreading the manuscript, Mr. Roger Tan Kay Chia, and Mr. Chew Hock See, and Ms. Kerh Geok Hong for their support.

**Electronic Supplementary Material**

# A ROBOTIC LEG INSPIRED FROM AN INSECT LEG

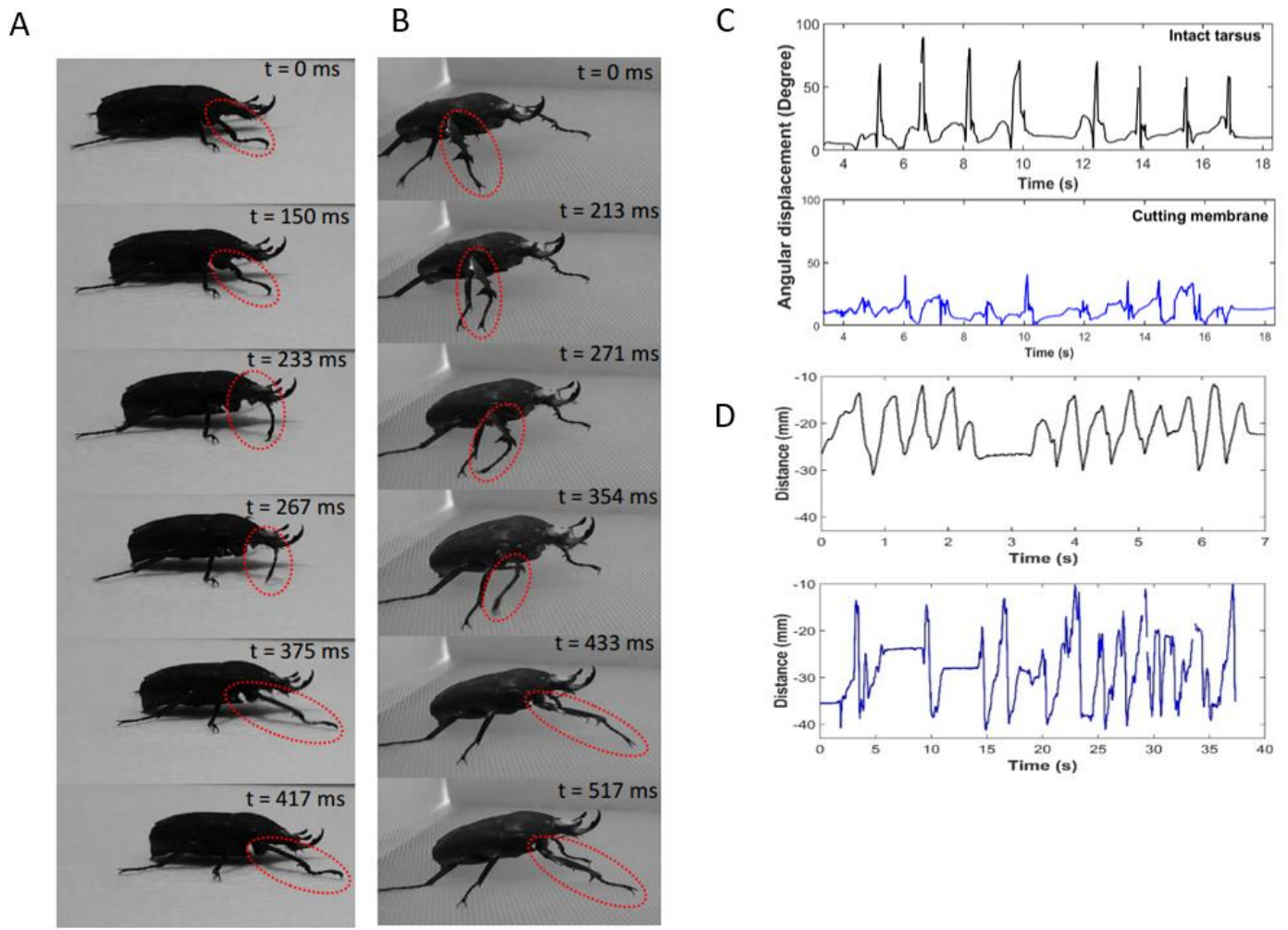

**Figure S1.** Observations and hypothesis on the tarsus and the bending ability of the tarsus. Tarsal deformation of the foreleg when the beetle is walking: frame-by-frame footage of the tarsus bending when walking (A) on the continuous surface (Styrofoam plate) and (B) on the flexible, mesh surface. (C) The result of cutting the membrane connecting the tarsomeres: Angular displacement comparison between the intact tarsus and the tarsus with the third tarsomere membrane cut. (D)The result of tubing the tarsus. The distance from the claw to the reference plane is put on the beetle body for intact tarsus and tubed tarsus.

Analysis of the cycle time and angular displacement:

The two samples t-test (0.05 significant level) was performed to check whether there is a significant difference between the means in two unrelated groups. The null hypothesis for the independent t-test is that the population means from the two unrelated groups are equal.

**Table S1:** Statistical analysis two-tailed t-test for the angular displacement when beetle walking on the mesh and the Styrofoam substrates (N = 5 beetles).

| Type of substrates | Mesh | Plate |
|---|---|---|
| Mean (Degree) | 55.7 | 27.7 |
| Standard deviation (Degree) | 4.4 | 5.4 |
| Hypothesized Mean Difference | 0 | |
| df | 8 | |
| P(T<=t) one-tail | 9.04E-06 | |
| P(T<=t) two-tail | 1.81E-05 | |

**Table S2:** Statistical analysis two-tailed t-test for the cycle time when beetle walking on the mesh and the Styrofoam substrates (N = 5 beetles).

| Type of substrates | Mesh | Plate |
|---|---|---|
| Mean (ms) | 446.1 | 406.6 |
| Standard deviation (ms) | 50.5 | 67.8 |
| Hypothesized Mean Difference | 0 | |
| df | 8 | |
| P(T<=t) one-tail | 0.1631 | |
| P(T<=t) two-tail | 0.3262 | |

**Table S3:** Statistical analysis two-tailed t-test for the angular displacement when beetle walking on the mesh with intact tarsus and with tarsus after cutting the membrane (N = 3 beetles).

| Type of substrates | Intact tarsus | Cutting membrane |
|---|---|---|
| Mean (Degree) | 55.9 | 25.5 |
| Standard deviation (Degree) | 2.3 | 1.6 |
| Hypothesized Mean Difference | 0 | |
| df | 4 | |
| P(T<=t) one-tail | 2.42E-05 | |
| P(T<=t) two-tail | 4.85E-05 | |

**Table S4:** Statistical analysis two-tailed t-test for the tarsal bending between two cases: intact tarsus of intact beetle (Fig. 5A, N = 5 beetles) and intact tarsus of the beetle that was cutting the membrane at another forelegs' tarsus (Fig.5B, N = 3 beetles)**.**

| Type of substrates | Intact beetle | Before cutting membrane |
|---|---|---|
| Mean (degree) | 55.7 | 55.9 |
| Standard deviation (degree) | 4.4 | 2.3 |
| Hypothesized Mean Difference | 0 | |
| df | 6 | |
| P(T<=t) one-tail | 0.4418 | |
| P(T<=t) two-tail | 0.8835 | |

**Table S5:** Statistical analysis two-tailed t-test for the cycle time when beetle walking on the mesh with intact tarsus and with tubed tarsus (N = 5 beetles).

| Type of substrates | Intact tarsus | Tubed tarsus |
|---|---|---|
| Mean (ms) | 446.1 | 1594.4 |
| Standard deviation (ms) | 50.5 | 142.5 |
| Hypothesized Mean Difference | 0 | |
| df | 8 | |
| P(T<=t) one-tail | 7.33E-08 | |
| P(T<=t) two-tail | 1.47E-07 | |

**Table S6:** Statistical analysis two-tailed t-test for the cycle time when beetle walking on the mesh with intact tarsus and with releasing tarsus after removing the tube in tubing experiment (N = 5 beetles).

| Type of substrates | Intact tarsus | Releasing tarsus |
|---|---|---|
| Mean (ms) | 446.8 | 443.36 |
| Standard deviation (ms) | 50.5 | 48.4 |
| Hypothesized Mean Difference | 0 | |
| df | 48 | |
| P(T<=t) one-tail | 0.4664 | |
| P(T<=t) two-tail | 0.9328 | |